# Understanding the Principles of Recursive Neural Networks: A Generative Approach to Tackle Model Complexity

Alejandro Chinea

Departamento de Física Fundamental, Facultad de Ciencias UNED,
Paseo Senda del Rey nº9, 28040-Madrid - Spain

**Abstract.** Recursive Neural Networks are non-linear adaptive models that are able to learn deep structured information. However, these models have not yet been broadly accepted. This fact is mainly due to its inherent complexity. In particular, not only for being extremely complex information processing models, but also because of a computational expensive learning phase. The most popular training method for these models is back-propagation through the structure. This algorithm has been revealed not to be the most appropriate for structured processing due to problems of convergence, while more sophisticated training methods enhance the speed of convergence at the expense of increasing significantly the computational cost. In this paper, we firstly perform an analysis of the underlying principles behind these models aimed at understanding their computational power. Secondly, we propose an approximate second order stochastic learning algorithm. The proposed algorithm dynamically adapts the learning rate throughout the training phase of the network without incurring excessively expensive computational effort. The algorithm operates in both on-line and batch modes. Furthermore, the resulting learning scheme is robust against the vanishing gradients problem. The advantages of the proposed algorithm are demonstrated with a real-world application example.

## 1 Introduction

Recursive neural networks were introduced last decade [1],[2] as promising machine learning models for processing data from structured domains (i.e.: protein topologies, HTML web pages, DNA regulatory networks, parse trees in natural language processing, and image analysis amongst others). These computational models are suited for both classification and regression problems being capable of solving supervised and non-supervised learning tasks. One of the goals behind this approach was to fill the existing gap between symbolic and sub-symbolic processing models. Specifically, to develop computational schemes able to combine numerical and symbolic information in the same model. Moreover, the principal advantage associated to them was their ability to work with patterns of information of different sizes and topologies (i.e. trees or graphs) as opposed to feature-based approaches where the information pertinent to the problem is encoded using fixed-size vectors. Graphs are more flexible data structures than vectors since they may consist of an arbitrary number of nodes and edges, while vectors are constrained to a predefined length which has to be preserved by all patterns composing the training set. So far, these models have found applications mainly in bio-informatics [3], [4] although they have also been applied to image analysis [5] and natural language processing tasks [6] between others.

However, despite the initial interest motivated by these models their development is still in its infancy. This fact can be explained in part because the inherent restrictions associated to the first models where the recursive paradigm was limited to work only with acyclic structures under causality and stationary assumptions, something too restrictive for many real-world problems. Another limiting factor to be considered is that structured domains possess very little mathematical structure. Specifically, basic mathematical operations such us computing the sum or the covariance of two graph objects are not available. Furthermore, it is also important to note that nowadays, early research problems like learning generic mappings between two structured domains (e.g.: IO-isomorphic and non IO-isomorphic structured transductions) still remains as challenging open research problems.

Although some advances have been recently reported regarding not only the recursive processing of cyclic structures [7],[8] but the contextual processing of information (i.e. recursive models breaking the causality hypothesis) [9],[10], followed by some basic proposals on generating structured outputs [11],[12], from a practical point of view, the intrinsic complexity of these models together with a computationally hard learning phase has strongly limited the interest of the research community on this kind of models. It is important to note that learning in structured domains has been traditionally considered a very difficult task. Furthermore, it has been recently pointed out [13] that two important future challenges for these models will rely on the design of efficient learning schemes, and tackling appropriately theoretical problems as learning structural transductions or structure inference as occur in various machine learning areas such as the inference of protein structures or parse trees.

In this paper we present an approximate stochastic second order training algorithm aimed to overcome the complexity of the training phase associated to these models. The main advantage of this training method is that is robust against the vanishing gradients problem. Furthermore, the resulting scheme leads to an algorithm which achieves an optimal trade-off between speed of convergence and the required computational effort. In addition, this paper puts also the emphasis on the analysis of the underlying principles of the computational model associated to recursive neural networks. This analysis will permit us to better understand their computational power. The rest of the paper is organized as follows: In the next section we analyze in detail the principles behind the computational model implemented by recursive neural networks. Section 3, is devoted to show the background of the approximate second order stochastic algorithm. In section 4, experimental results are provided comparing the proposed algorithm with existing approaches. Finally, concluding remarks are outlined in section 5.

## 2 Implications of the Recursive Neural Model

### 2.1 Definitions and Notations

A graph *U* is a pair *(V,E)*, where *V* is the set of nodes and E represents the set of edges. Given a graph *U* and a vertex *v* ∈ *V*, *pa[v]* is the set of parents of v, while *ch[v]* represents the set of its children. The in-degree of *v* is the cardinality of *pa[v]*, while its out-degree is the cardinality of *ch[v]*. Under the recursive model the patterns of information are labeled graphs. Specifically, the graph nodes contain a set of domain variables characterized by a vector of real and categorical variables. Furthermore, each node encodes a fragment of information that is believed to play an important role in the task at hand. The presence of a branch *(v,w)* between two nodes explicitly models a logical relationship between the fragments of information represented by nodes *v* and *w*.

The recursive neural network model is composed of a state transition function f and an output function g (see figure 1). These functions are usually implemented by multi-layer perceptron networks. The standard model is suited to process directed positional acyclic graphs with a super-source node. Furthermore, they implement deterministic IO-isomorphic transductions based on the following recursive state representation:

$$a(v) = f\big(a(ch[v]), I(v), v, W_f\big)$$
$$y(v) = g\big(a(v), v, W_g\big) \quad (1)$$

In expression (1) $W_f$ and $W_g$ represent the synaptic weights (model parameters) of networks f and g respectively. In order to process a graph *U* the state transition network is unfolded through the structure of the input graph leading to the encoding network. This unfolding procedure is followed in both learning and recall phases of the neural network. The resulting network has the same structure of the input graph, while nodes are replaced by copies of the state transition network and a copy of the output network is inserted at the super-source. Afterwards, a feed-forward computation is carried out on the encoding network. More specifically, at each node *v* of the graph, the state *a(v)* is computed by the transition network as a function of the input label *I(v)* and the state of its children (first equation of expression (1)) with:

$$a(ch[v]) = [a(ch_1[v]), a(ch_2[v]), \ldots, a(ch_o[v])] \quad (2)$$

In expression (2) the index *o* stands for the maximum out-degree of node *v*. The base case for the recursion is a(nil) = $a_0$ which correspond to a frontier state (e.g. if node *v* lacks of its *i*-th child). At the super-source (node s) the output is computed as *y = g(a(s))*. It is important to note that the standard model of recursive neural networks implements transductions consisting on the mapping of input structures U into an output structure Y which is always made of a single node (the output is not structured).

## 2.2 Generative Principles of Intelligence

Generally speaking, the notion of intelligence cannot be conceived without the existence of two fundamental principles [14]: The maximization of transfer and the recoverability principle. Basically, the first principle states that whenever it is possible an intelligent system builds knowledge by maximizing the transfer of previously acquired knowledge. Specifically, more complex structures are the result of assembling previously learnt or memorized structures (e.g. to transfer actions used in previous situations to handle new situations). However, this mechanism of transfer must be understood as a gestalt procedure (the resulting structures being more than a simply combination of previously learnt structures). In addition, the recoverability principle is associated to the concept of memory and inference. It states that a system displaying intelligence must be able to recover itself from its actions. Specifically, an intelligent system must be able to infer causes from its own current state in order to identify what it failed or succeed something not possible without the existence of memory.

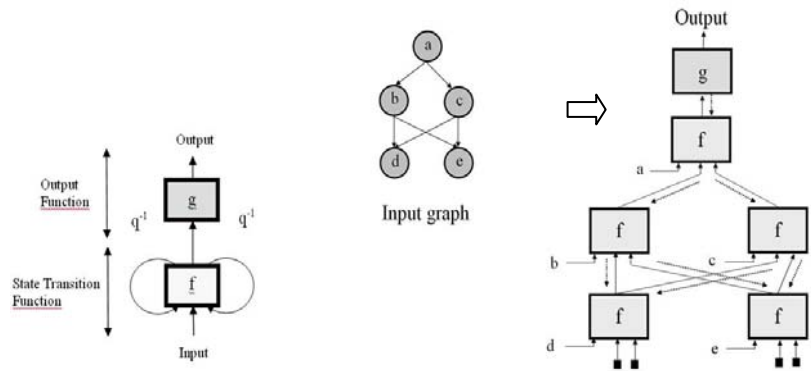

**Fig. 1.** Block diagram of the Recursive neural network model (left side of the figure). The right side of the picture depicts the unfolding procedure.

It is has been recently shown [15] that the human information processing system follows these two generative principles. Specifically, the perception system organizes the world by using cohesive structures. Furthermore, such a structure is the result of a hierarchically organized information processing system [16],[17] that generates structure by correlating the information processed at the different levels of its hierarchy. As a result of this process, world complexity is turned into understanding by finding and assigning structure. This mechanism, amongst many other things, permits us to relate objects of different kinds. In addition, the nature of the perceived structures and their relationships is also linked to the context under consideration. For instance, the relationships between objects can be causal in a temporal context, geometrical in pattern recognition problems (e.g. the shapes appearing in an image) or topological in bio-informatics problems (e.g. chemical compounds, protein structures etc).

### 2.3 Recursive Networks as Generative Models

Under the recursive neural networks framework the perceived structure of a problem is captured and expressed by using graphical models. In particular, the patterns used for the learning and recall phases not only encode the fragments of information (e.g.: information that can be characterized by specific attributes that are quantifiable and/or measurable) which play an important role in the machine learning problem but also the logical relationships between them. The nature of such relations is determined by the application context and attempts to explicitly model the logical correlations between fragments of information. For instance, in a temporal domain the fragments of information are events and the co-occurrence of two or more events is interpreted as an existing or possible correlation between them. Therefore, this information encoding procedure contains more knowledge rather than if such pieces of information were considered in isolation. It is important to note that the notion of information content is strongly connected to the notion of structure. Indeed, the fact of building more complex structures from more basic ones is reflecting the first of the two generative principles related to the concept of intelligence resulting in a gain not only in information representation but in information content.

On the other hand, the computational scheme imposed by the recursive state equation (1) leads to a sequential propagation of information which follows a reversed topological sort of the input graph during the recall phase of the network (hidden states are updated starting from leaves toward the super-source node). In addition, this flow of information is bidirectional during the learning phase. The main consequence of this message passing procedure is that the computational model of recursive networks can be viewed as an inference system that learns the hidden dependencies explicitly encoded within the structural patterns used for the training phase of the network. Furthermore, as stated in [18] recursive neural networks can be viewed as limits, both in distribution and probability, of Bayesian networks with local conditional distributions. Therefore, they implement in a simplified form the notion of recoverability.

## 3 Reducing Complexity of the Training Phase

### 3.1 Theoretical Background

The concept of learning in neural networks is associated to the minimization of some error function *E(W)* by changing network parameters *W*. In the case of recursive networks the learning phase consists in finding the appropriate model parameters for implementing the state transition network f and the output network g with regards to the given task and data. Without a loss of generality let us suppose that we express the parameters of function f and g in a vector $W=[w_1,w_2,w_3,....,w_m]$. A perturbation of the error function around some point of the model parameters which can be written as follows: $E(W+\Delta W) = E(w_1 + \Delta w_1, w_2 + \Delta w_2, ..., w_m + \Delta w_m)$. Considering the Taylor expansion of the error function around the perturbation $\Delta W$ we obtain:

$$E(W + \Delta W) = E(W) + \sum_{i=1}^{m}\frac{\partial E(W)}{\partial w_i}\Delta w_i + \frac{1}{2}\sum_{i=1}^{m}\frac{\partial^2 E(W)}{\partial w_i^2}(\Delta w_i)^2 + \sum_{i<j}\frac{\partial^2 E(W)}{\partial w_i \partial w_j}\Delta w_i \Delta w_j + \frac{1}{6}\sum_{i=1}^{m}\frac{\partial^3 E(W)}{\partial w_i^3}(\Delta w_i)^3 + ....... \quad (3)$$

The training phase consists roughly in updating the model parameters after the presentation of a single training pattern (on-line mode) or batches of the training set (off-line or batch mode). Each update of model parameters can be viewed as perturbations (e.g. noise) around the current point given by the m dimensional vector of model parameters. Let us assume a given sequence of N disturbance vectors $\Delta W$. Ignoring third and higher order terms in expression (3), the expectation of the error $<E(W)>$ can be expressed as:

$$\langle E(W) \rangle \cong \frac{1}{N}\sum_{n=1}^{N}E(W + \Delta W^n) \quad (4)$$

$$\langle E(W) \rangle = E(W) + \sum_{i=1}^{m}\frac{\partial E(W)}{\partial w_i}\frac{1}{N}\sum_{n=1}^{N}\Delta w_i^n + \frac{1}{2}\sum_{i=1}^{m}\frac{\partial^2 E(W)}{\partial w_i^2}\frac{1}{N}\sum_{n=1}^{N}(\Delta w_i^n)^2 + \sum_{i<j}\frac{\partial^2 E(W)}{\partial w_i \partial w_j}\frac{1}{N}\sum_{n=1}^{N}\Delta w_i^n \Delta w_j^n$$

Rearranging the previous expression we obtain a series expansion of the expectation of the error in terms of the moments of the random perturbations:

$$\langle E(W) \rangle \cong E(W) + \sum_{i=1}^{m}mean(\Delta w_i)\frac{\partial E(W)}{\partial w_i} + \frac{1}{2}\sum_{i=1}^{m}var(\Delta w_i)\frac{\partial^2 E(W)}{\partial w_i^2} + \sum_{i<j}cov(\Delta w_i, \Delta w_j)\frac{\partial^2 E(W)}{\partial w_i \partial w_j} \quad (5)$$

In addition, the weight increment associated to the gradient descent rule is $\Delta w_i = -\eta g_i$. The third term of expression (4) concerning the covariance can be ignored supposing that the elements of the disturbance vectors are uncorrelated over the index n. This is a plausible hypothesis given that patterns are presented randomly to the network during the learning phase. Moreover, close to a local minimum we can assume that $mean(\Delta w_i) \approx 0$ (the analysis of this approximation is omitted for brevity, but they show that its effects can be neglected). Taking into account these considerations the expectation of the error is then given by:

$$\langle E(W) \rangle \cong E(W) + \frac{1}{2}\sum_{i=1}^{m}\sigma^2(\Delta w_i)\frac{\partial E(W)}{\partial w_i^2} = E(W) + \frac{\eta^2}{2}\sum_{i=1}^{m}\sigma^2(g_i)\frac{\partial^2 E(W)}{\partial w_i^2} \quad (6)$$

From equation (5) it is easy to deduce that the expected value of the error increases as the variance (represented by the symbol $\sigma^2$) of the disturbance vectors (gradients) increases. This observation suggests that the error function should be strongly penalized around such weights. Therefore, to cancel the noise term in the expected value of the error function the gradient descent rule must be changed to $\Delta w_i = -\eta g_i/\sigma(g_i)$. This normalization is known as vario-eta and was proposed [19] for static networks. Specifically, the learning rate is renormalized by the stochasticity of the error signals. In this line, it is important to note that the error signals contain less and less information at the extent they pass through a long sequence of layers, due to the shrinking procedure carried out by the non-linear (hyperbolic tangent or sigmoidal functions) network units. This is a well-known problem [20] that makes very difficult to find long-term dependencies that could eventually appear encoded within the structural patterns of the training set. For the case of recursive neural networks this problem is even worst due to the unfolding procedure (error signals can traverse many replicas of the same network). However, this normalization procedure avoids the gradients vanishing thanks to the scaling of network weights.

### 3.2 Algorithm Description

The whole algorithmic description is provided in figure 2. By inspection of the pseudo-code, the algorithm proceeds as follows: after the initialization of algorithm parameters (lines 1 and 2), the algorithm enters in two nested loops: The external or control loop is in charge of monitoring the performance of the algorithm.

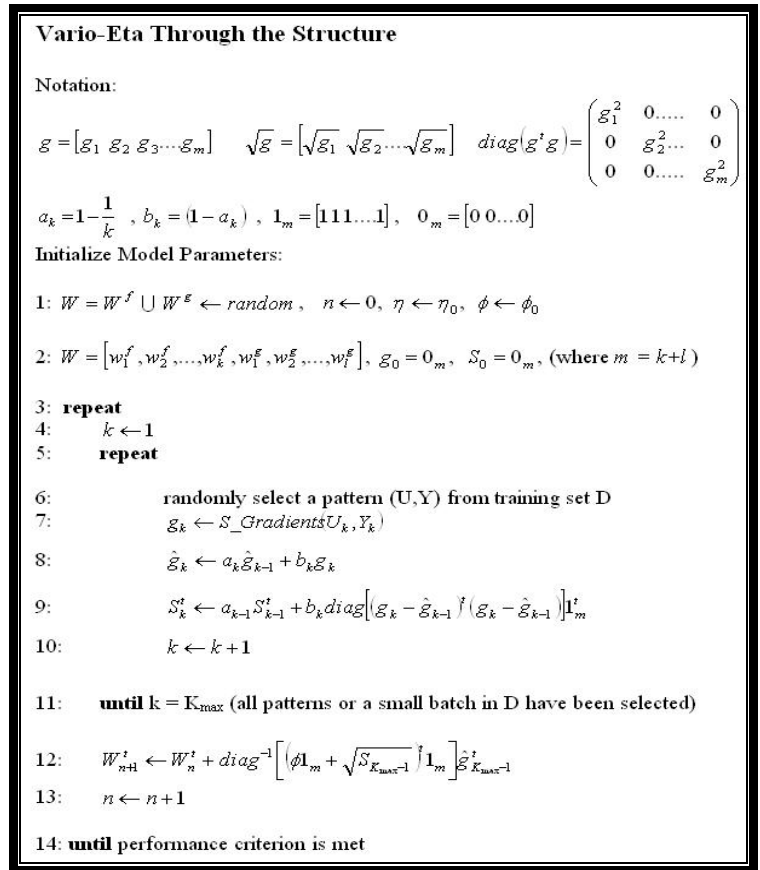

**Fig. 2.** Pseudo-code of the proposed approximated second order stochastic learning algorithm.

In addition, it also performs the update of model parameters (line 12) following the derived learning rule ($\Delta w_i = -\eta g_i/\sigma(g_i)$). The constant $\phi$ is summed to the standard deviation of the error gradients for avoiding eventual numerical problems.

Figure 3 shows the pseudo-code of the function S_Gradients (S stands for Structure). This function is in charge of computing the first derivatives of the error function with respect to model parameters. This function takes as arguments a structural pattern U together with its category Y and returns the error gradient vector. The details of the notation used can be found in [22].

```
S_Gradients(U,Y)
Choose a reverse topological ordering ≻ on U
for each node u ∈ U obeying the ordering ≻ do
    u.a ← f(ch[u], I(u))
endfor
for each u ∈ U do
u.δ^f ← 0
G^f ← 0, G^g ← 0
Choose a topological ordering on ≺ U
for each node u ∈ U obeying the ordering ≺ do
    δ^g ← y(u.a) − t(u)
    G^g ← G^g + J_w^g(u.a)δ^g
    u.δ^f ← u.δ^f + J_x^g(u.a)δ^g
    G^f ← G^f + J_w^f(a(ch[u]), I(u)) u.δ^f
    for z ∈ ch[u] do
        r ← ord(z, ch[u])
        z.δ^f ← z.δ^f + J_{x(r)}^f(a(ch[u]), I(u)) u.δ^f
    endfor
endfor
return G = G^f ∪ G^g
```

**Fig. 3.** Pseudo-code of the function in charge of computing the first derivatives of the error with respect to model parameters.

The internal loop (lines 5 up to 11) operates recursively in order to obtain the variance (line 9) of the error gradients (line 7). Similarly, the mean value of the gradients is computed recursively in line 8. It is important to note that the algorithm can operate in both batch mode $k_{max} = \|D\|$ (number of training patterns) or in on-line mode $k_{max} \ll \|D\|$ just by selecting appropriately the value of $k_{max}$. Finally, it is important to note that the proposed algorithm can be easily adapted to other extensions of the recursive model like contextual models or graph neural networks.

### 3.3 Preliminary Complexity Analysis

From a computational point of view, the proposed algorithm scales $O(W)$ in terms of memory storage requirements, where $W = W_f \cup W_g$ is the number of parameters of the model. In addition, the computational cost scales roughly as $O(NW)$ where N is the number of patterns in the data set. In this line, it must be noted that quasi-Newton methods that builds iterative approximations of the inverse Hessian lead in general, to algorithms with overall computational cost of $O(NW^2)$ and memory storage requirements of $O(W^2)$. Similarly, conjugate-gradients method achieves a memory storage of $O(W)$ at the same computational cost of a quasi-Newton method. In addition, it is important to note that a rigorous complexity analysis would require a careful study of the statistical distribution of the gradient errors throughout the optimization procedure followed by an analysis of the generating functions associated to the recursions (lines 8 and 9 of pseudo-code of figure 2).

Therefore, taking into account that the proposed learning rule behaves like a stochastic approximation of a quasi-Newton method, the proposed algorithm achieves a good trade-off in terms of memory storage and computational complexity.

## 4 Experimental Results

The performance of the algorithm was tested for the problem exposed in [21]. Specifically, this application comes from the intelligent transportation research field and consists on the development of an advanced intersection safety system. The ultimate goal is to provide appropriate warnings to the driver to avoid fatal collisions. For this task, the structural patterns are trees ranging from one up to sixteen levels depth encoding temporal situations at road intersections. Generally speaking, a road intersection situation is composed of a set of dynamic (eg: vehicles, pedestrians, traffic lights, etc) and static entities (eg: trees, bushes, road signs, etc) interacting during a variable time frame. For this application, the pattern set is composed of 4000 structural patterns where approximately half of them representing highly-risky situations (e.g.: situations leading to collisions). The dynamic aspects of road intersections are encoded within the topology of the trees while static aspects are encoded within the label space of the tree-graphs. This task provides an illustrative example of how an extremely complex problem is modeled using the recursive paradigm.

Figure 4 provides a comparison of Back-propagation through the structure (Bpts), a quasi-Newton through the structure algorithm [21] (Qnts) and the proposed algorithm (Vets) running in batch mode. Graphics depict the result of averaging 10 simulations for two different network architectures. Specifically, each simulation consist on running a training phase of 20 epochs for each algorithm starting from identical weight initialization conditions. Afterwards, the resulting error values are normalized for the three algorithms in the interval [0,1] (the value that is mapped to 0 is the minimum value reached at epoch 20 by the best performing algorithm).

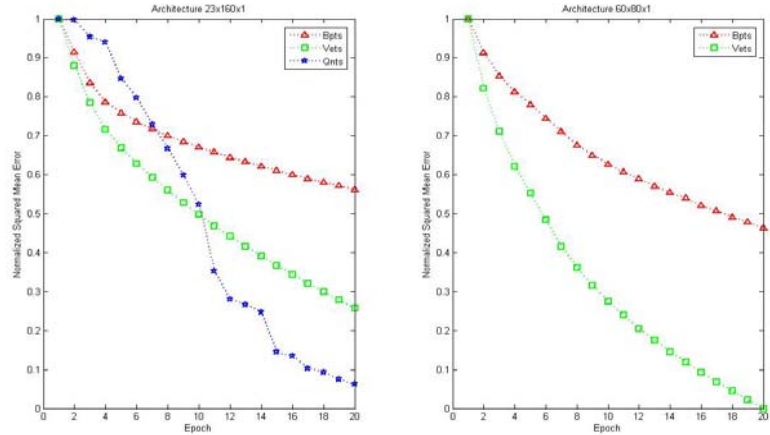

**Fig. 4.** Simulation results for the proposed stochastic algorithm (Vets), back-propagation through the structure (Bpts) and a quasi-Newton algorithm (Qnts).

Finally, the normalized error values are averaged over the 10 simulations. The left side of the figure shows the averaged results for an architecture of 23x160x1 (23 units implementing the state transition function and 161 units for the output function). Similarly, the right side of the picture shows the results for a 60x80x1 architecture. Due to the prohibitive memory requirements of the quasi-Newton algorithm (Hessian matrix contains more than $10^7$ elements) the comparison was only possible with the Bpts algorithm for this network architecture.

Therefore, although further experimentation must be carried out the proposed stochastic algorithm provides a good trade-off between the memory storage requirements and algorithm complexity.

## 5 Conclusions

In this paper we have described the principles behind the recursive neural network model. It was shown that associated with any given problem, the information content presents a certain geometry that these models can attempt to exploit. Furthermore, the fact of using structured representations of information is translated into a substantial gain in information content. In addition, in order to tackle the inherent complexity of these models a stochastic learning algorithm was also described. The proposed algorithm is able to achieve a good trade-off between speed of convergence and the computational effort required by setting the local learning rate for each weight inversely proportional to the standard deviation of its stochastic gradient. The scaling properties of the algorithm make it suitable for the computational requirements of the recursive model. Furthermore, the proposed learning scheme can be easily adapted to other recursive models such as contextual models or graph neural networks. The computer simulations demonstrated the efficiency of the algorithm for a practical learning task.